# Extended Histogram-based Outlier Score (EHBOS)

**Tanvir Islam**
Okta
Bellevue, WA, USA
tanvir.islam@okta.com

## Abstract

Histogram-Based Outlier Score (HBOS) is a widely used outlier or anomaly detection method known for its computational efficiency and simplicity. However, its assumption of feature independence limits its ability to detect anomalies in datasets where interactions between features are critical. In this paper, we propose the Extended Histogram-Based Outlier Score (EHBOS), which enhances HBOS by incorporating two-dimensional histograms to capture dependencies between feature pairs. This extension allows EHBOS to identify contextual and dependency-driven anomalies that HBOS fails to detect. We evaluate EHBOS on 17 benchmark datasets, demonstrating its effectiveness and robustness across diverse anomaly detection scenarios. EHBOS outperforms HBOS on several datasets, particularly those where feature interactions are critical in defining the anomaly structure, achieving notable improvements in ROC AUC. These results highlight that EHBOS can be a valuable extension to HBOS, with the ability to model complex feature dependencies. EHBOS offers a powerful new tool for anomaly detection, particularly in datasets where contextual or relational anomalies play a significant role.

## 1 Introduction

Outlier detection, also referred to as anomaly detection, plays a critical role in numerous real-world applications, including fraud detection (Anbarasi & Dhivya, 2017), cyber security (Benjelloun et al., 2019), financial analytics (Hilal et al., 2022), healthcare monitoring (Bauder & Khoshgoftaar, 2016), retail behavior analysis (Yoseph et al., 2019), and more. Outliers or anomalies are generally the instances that deviate significantly from the norm or expected pattern (Hodge & Austin, 2004; Zhao et al., 2019). The ability to identify these instances is crucial for mitigating risks, optimizing processes, and improving decision-making. However, achieving efficient, accurate, and interpretable anomaly detection, particularly in large and high-dimensional datasets, remains a significant challenge.

Outlier detection methodologies generally span statistical, distance-based, density-based, model-based, neural network-driven, and histogram-based paradigms. Statistical techniques, such as Z-score and Mahalanobis distance, rely on distributional assumptions to flag deviations, though they falter when such assumptions are violated (Rousseeuw & Hubert, 2011). Distance-based methods such as kNN identify outliers by measuring the distance between data points; points that are far away from their nearest neighbors are considered outliers (Angiulli & Pizzuti, 2002; Chen et al., 2010). Density-aware methods like Local Outlier Factor (LOF) identify outliers as instances whose local density is significantly lower than that of their neighbors (Breunig et al., 2000). Model-based frameworks, such as One-Class SVM (Schölkopf et al., 2001) and Isolation Forest (Liu et al., 2008), learn decision boundaries or partitions to separate anomalies. Neural network methods, notably autoencoders (Sakurada & Yairi, 2014), excel in high-dimensional spaces by reconstructing inputs and flagging outliers via reconstruction error; they capture complex interactions but require significant computational resources and risk overfitting. Histogram-based techniques, exemplified by HBOS (Histogram-Based Outlier Score), offer a computationally efficient alternative by constructing univariate histograms per feature, aggregating inverse likelihoods to score anomalies—ideal for large datasets but limited in capturing multivariate interactions (Goldstein & Dengel, 2012).

Among the many methods proposed for anomaly detection, the Histogram-based Outlier Score (HBOS) has emerged as a popular approach due to its computational efficiency and simplicity. By leveraging histograms to model the distribution of feature values, HBOS offers a scalable and interpretable framework for detecting outliers in multidimensional data. Despite its strengths, HBOS suffers limitations due to its univariate approach. HBOS analyzes each feature independently. This means it can effectively detect outliers that have unusual values in a single feature, but it may miss outliers that are unusual due to a specific combination of feature values. It cannot capture complex dependencies between features.

To address this limitation, we propose the Extended Histogram-based Outlier Score (EHBOS), a novel enhancement to the HBOS framework. EHBOS builds on the foundation of HBOS while introducing critical extensions that improve its accuracy, robustness, and applicability in certain scenarios. EHBOS incorporates mechanisms to account for feature interactions, and introduces new strategies for modeling complex datasets. Specifically, it incorporates pairwise dependencies among features.

It is worth mentioning that several subsequent works have explored different avenues to enhance anomaly detection in this area.  Some of which directly extend or are inspired by HBOS's principles, while others offer alternative approaches to address similar challenges. For instance, Aguilera-Martos et al. (2023) proposed Multi-step Histogram-based Outlier Scores, which enhance HBOS by incorporating a multi-step approach to better identify anomalies in engineering datasets. Another notable extension is Subspace Histograms for Outlier Detection by Sathe and Aggarwal (2018), which addresses the high-dimensional challenge by applying HBOS principles within various subspaces, thus capturing more intricate outlier patterns. Furthermore, Aryal et al. (2019) introduced an Improved Histogram-Based Anomaly Detector with Extended Principal Component Features, integrating principal component analysis to enhance the detector's capability in handling correlated features. These advancements highlight the ongoing research efforts to overcome the inherent limitations of the original HBOS and underscore the need for further improvements, which our proposed EHBOS aims to provide.

The remainder of this paper is organized as follows: Section 2 discusses on the limitations of HBOS and positions EHBOS as a necessary extension. Section 3 introduces the EHBOS methodology in detail, including its extensions and theoretical underpinnings. Section 4 presents experimental results on benchmark datasets, comparing EHBOS to the HBOS. Finally, Section 5 concludes with a discussion of key findings, and directions for future research.

## 2 HBOS Limitations

Histogram-Based Outlier Score (HBOS) is a computationally efficient anomaly detection method that estimates outlier scores based on feature-wise histograms. However, its core assumption of feature independence fundamentally limits its ability to handle datasets where interactions between features are critical for identifying anomalies. To illustrate this limitation, we consider two synthetic simple datasets and examine the behavior of HBOS outlier scores.

Figure 1 presents two example datasets, while the corresponding HBOS scores are shown in Figure 2. In the first example, a dataset is composed of two well-separated clusters of normal data, with two localized outliers positioned far from the clusters. Although these outliers are clearly anomalous when the joint distribution of the features is considered, their individual feature values do not significantly deviate from the marginal distributions of the normal clusters. Consequently, HBOS, which independently evaluates each feature's histogram, assigns these outliers low outlier scores. This is evident in the corresponding HBOS outlier scores, where the outliers are indistinguishable from normal points. This behavior highlights HBOS's inability to detect contextual anomalies that depend on relationships between features.

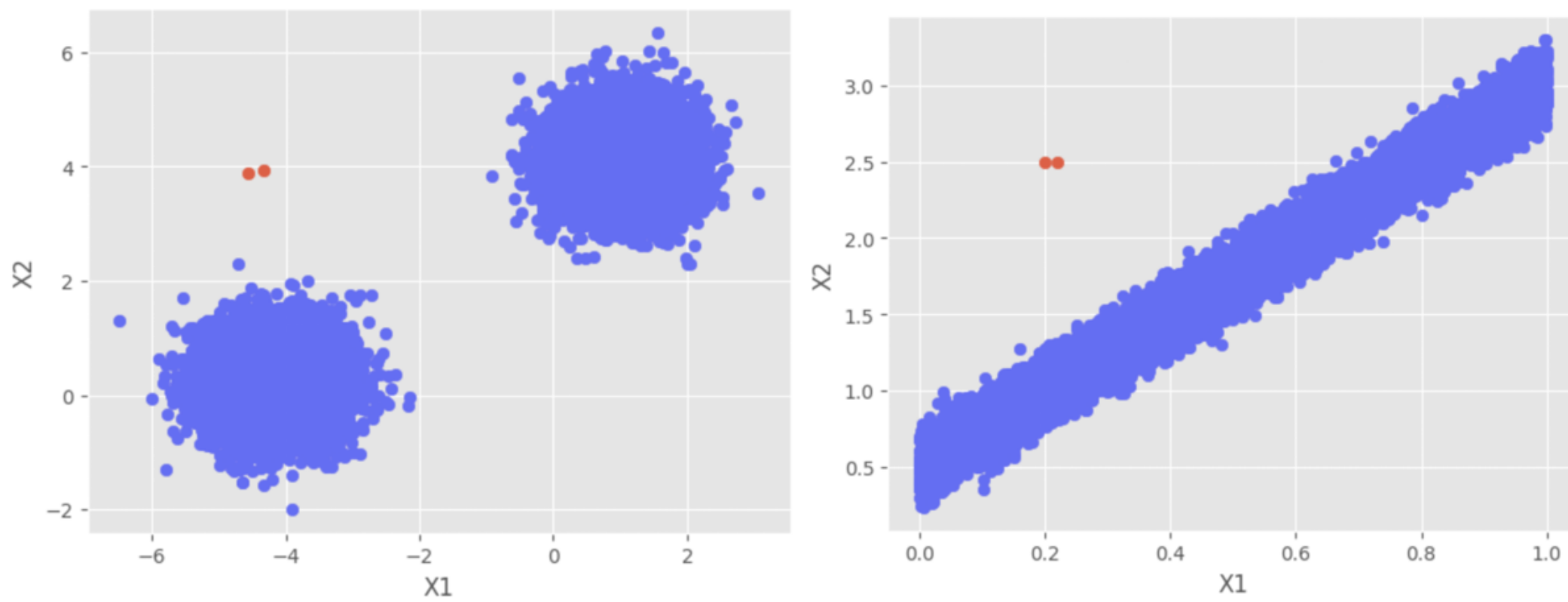

**Figure 1:** Examples of two datasets illustrating the limitations of HBOS. (Left) A clustered dataset with two localized outliers that are anomalous in the joint feature space but not in individual marginal distributions. (Right) A regression-based dataset where normal data follows a strong linear dependency, and two outliers deviate from this relationship while remaining within typical marginal ranges.

The second example considers a regression-based dataset in which the normal data follows a strong linear dependency between two features. Two outliers, while consistent with the marginal distributions of the individual features, deviate significantly from the linear relationship governing most of the data. When applied to this dataset, HBOS fails to recognize these outliers, as it does not capture the underlying feature dependency. Again, the HBOS scores demonstrates that the anomalies receive low scores, similar to the majority of the normal points. This failure arises from HBOS treating each feature independently, ignoring the critical interactions that define the structure of the normal data.

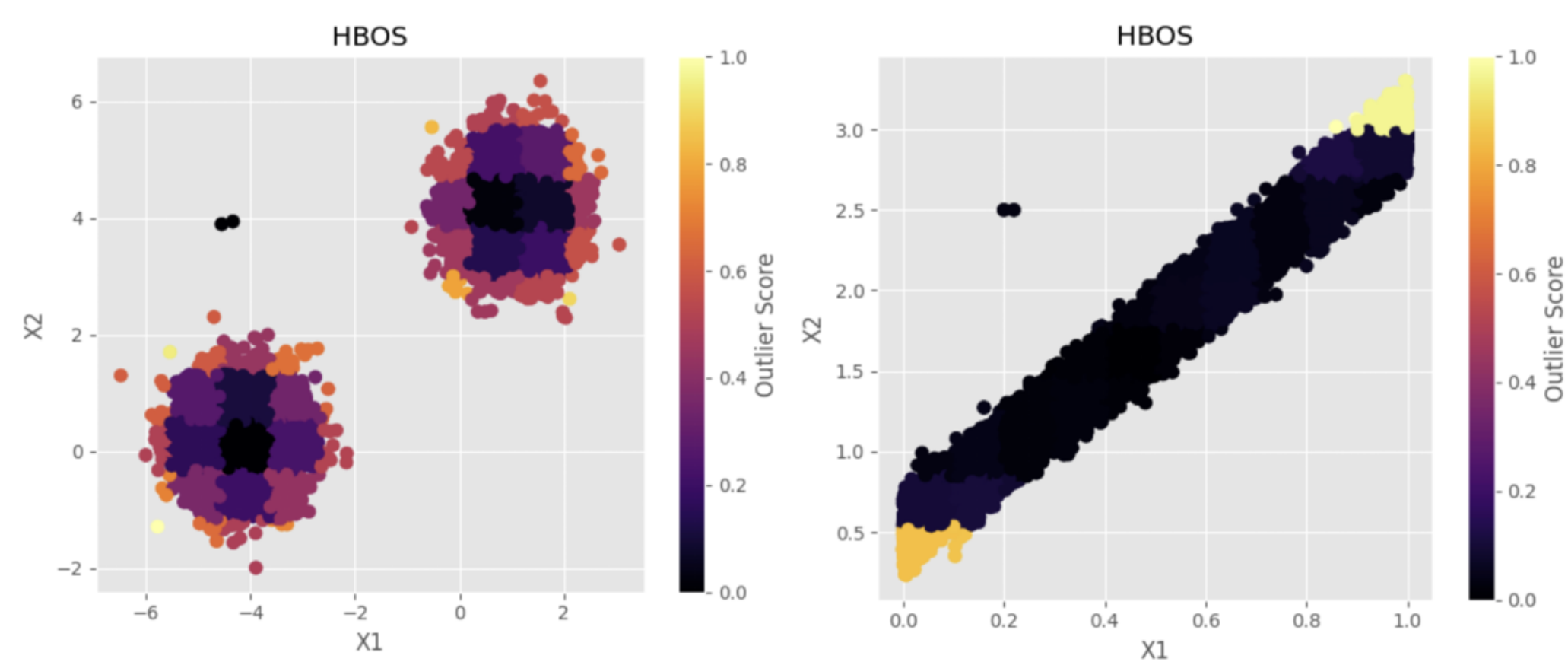

**Figure 2:** Histogram-based outlier scores assigned by HBOS for the datasets in Figure 1. Due to its independent feature analysis, HBOS fails to differentiate the outliers from normal points, resulting in low anomaly scores. This demonstrates HBOS's inability to detect anomalies arising from feature interactions.

In both examples, HBOS demonstrates its inability to detect anomalies that arise from deviations in feature relationships rather than individual feature values. These limitations are especially problematic in real-world datasets where interactions between features can be critical, as HBOS relies solely on univariate histograms and cannot account for such dependencies. This fundamental shortcoming motivates the need for extensions, such as EHBOS, that incorporate multi-dimensional feature interactions for more robust anomaly detection.

## 3 EHBOS

The Extended Histogram-Based Outlier Score (EHBOS) enhances the traditional Histogram-Based Outlier Score (HBOS) by combining one-dimensional (1D) and two-dimensional (2D) scores. The 1D score is derived directly from HBOS, leveraging independent histograms for each feature. The 2D score, in contrast, models pairwise interactions between features using joint histograms, capturing dependencies that 1D histograms cannot. By aggregating normalized 1D and 2D scores, EHBOS achieves improved detection of outliers in datasets with both independent and dependent features.

Let's assume, our dataset is represented as $X \in R^{n \times d}$, where n is the number of samples and d is the number of features. For each feature $j \in 1, \dots, d$, the one-dimensional density is estimated using histograms. The density for feature j is computed as:

$$h_j(x_{i,j}) = \frac{\text{count}(x_{i,j} \in \text{bin}_b)}{n \cdot \text{binwidth}_b}$$

where $x_{i,j}$ is the value of the j-th feature for sample i, $\text{count}(x_{i,j} \in \text{bin}_b)$ is the number of samples in bin b, and n is the total number of samples, $h_j(x_{i,j})$ represents the height of the bin (density estimate). The histogram is normalized so that its maximum height is 1.0 across all bins, ensuring equal weight for each feature in the final outlier score computation.

The one-dimensional outlier score for sample i is then defined as:

$$s_i^{(1D)} = \sum_{j=1}^{d} \log\left(\frac{1}{h_j(x_{i,j})}\right) = \sum_{j=1}^{d} -\log h_j\left(x_{i,j}\right)$$

To capture feature interactions, EHBOS computes pairwise densities for all feature pairs $(j, k)$, where $j, k \in 1, \dots, d$ and $j < k$. The bin height is estimated as:

$$h_{jk}(x_{i,j}, x_{i,k}) = \frac{\text{count}\left((x_{i,j}, x_{i,k}) \in \text{bin}_{b_{jk}}\right)}{n \cdot \left(\text{binarea}_{b_{jk}}\right)}$$

where $\text{count}\left((x_{i,j}, x_{i,k}) \in \text{bin}_{b_{jk}}\right)$ is the number of samples in $\text{bin}_{b_{jk}}$, and $\text{binarea}_{b_{jk}}$ is the area of the two-dimensional bin, $h_{j,k}(x_{i,j}, x_{i,k})$ is the height (density estimate) of the 2D bin.

The two-dimensional outlier score for sample i is given by:

$$s_i^{(2D)} = \sum_{j=1}^{d} \sum_{k=j+1}^{d} -\log h_{jk}\left(x_{i,j}, x_{i,k}\right)$$

The final EHBOS score is computed by aggregating the normalized one-dimensional and two-dimensional scores as:

$$s_i^{\text{EHBOS}} = \frac{s_i^{(1D)} + s_i^{(2D)}}{2}$$

A summarized version of the algorithm is presented here.

**EHBOS Algorithm**

**Input**: $X \in R^{n \times d}$ - input data with $n$ samples and $d$ features
**Output**: $s$ - outlier scores for all samples
Compute 1D histogram-based outlier scores using HBOS.
Normalize the 1D scores.
Initialize 2D outlier scores to zero.
for each feature pair $(j,\ k)$ where j $< k$ do
    Extract the two-dimensional feature subset.
    Compute 2D histogram-based outlier scores using HBOS.
    Accumulate the normalized 2D scores.
end for
Compute the final EHBOS outlier scores by averaging 1D and 2D scores.
return the final outlier scores.

It is worth noting that while HBOS offers an appealingly efficient approach to anomaly detection with a linear time complexity of $O(n \cdot d)$ (where $n$ is the number of samples and $d$ is the number of features) due to its reliance on independent 1D histograms, EHBOS inherently incurs a higher computational cost. By extending HBOS to incorporate 2D histograms that capture pairwise feature dependencies, EHBOS's complexity becomes $O(n \cdot d^2)$ when considering all unique feature pairs. This quadratic dependency on the number of features stems from the need to construct and evaluate a 2D histogram for each of the (2D) feature combinations. This increased complexity is a direct consequence of EHBOS's enhanced capability to detect anomalies that manifest through the relationships between features, a type of outlier often might be missed by purely univariate methods.

# 4 Results and Discussion

### 4.1 Simple Datasets

To further elucidate the benefits of EHBOS, we revisit the results on the two simple datasets discussed earlier: the local outliers in clusters dataset and the regression outliers dataset. As previously highlighted in the HBOS limitations section, HBOS fails to assign high outlier scores to the outliers in both cases due to its inability to capture feature dependencies. In contrast, EHBOS leverages two-dimensional histograms to model feature interactions, resulting in significantly higher outlier scores for the anomalies. Figure 3 presents these comparisons, where, for illustration purposes, we display only the 2D score $s_i^{(2D)}$ from EHBOS. In practice, the choice between using the 2D score alone or the averaged final score in EHBOS can be treated as a tunable hyperparameter.

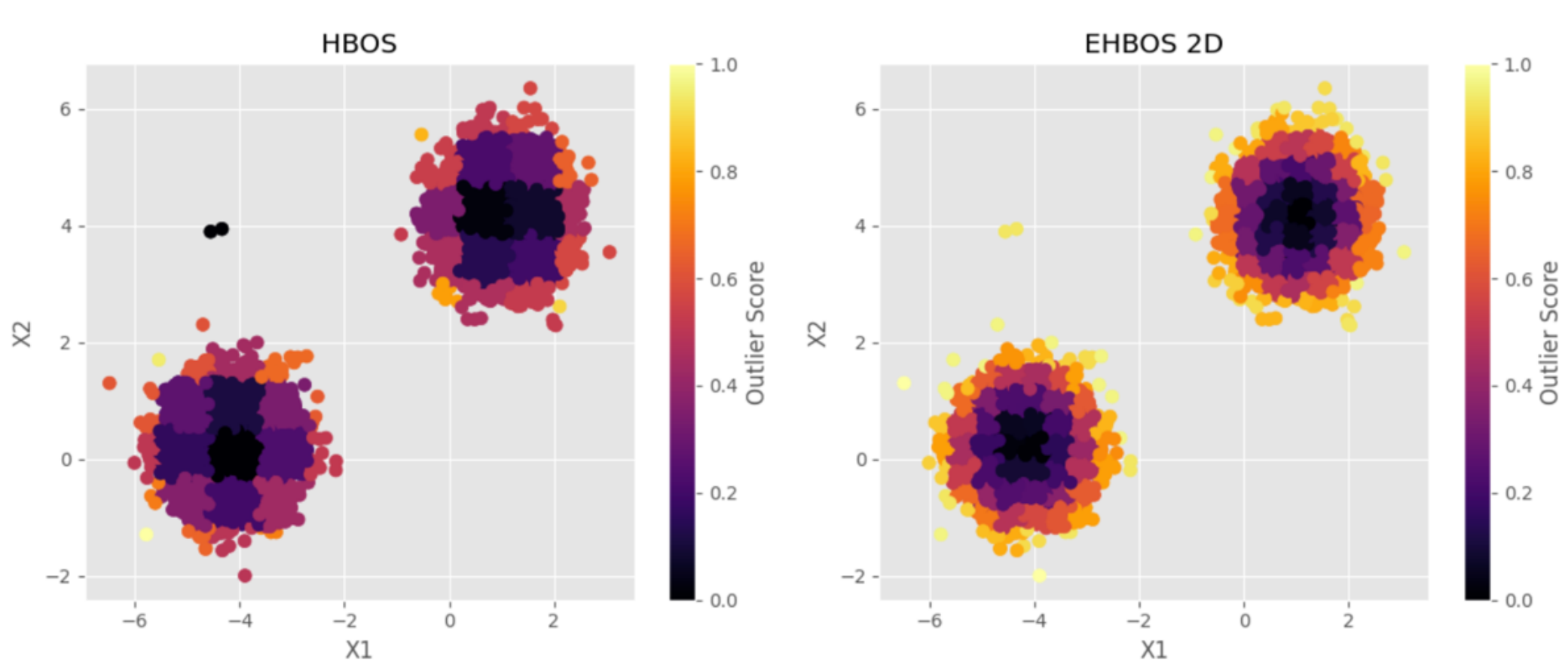

**Figure 3:** Comparison between HBOS and EHBOS (2D) outlier scores for the first example dataset.

In the local outliers dataset, EHBOS recognizes that the outliers deviate significantly from the joint distributions of the two features, even though their marginal distributions appear consistent with the normal clusters. This ability to model contextual dependencies enables EHBOS to assign high anomaly scores to these points, as shown in the corresponding scatter plot and colormap plot of EHBOS scores (Figure 3). Similarly, in the regression outliers dataset, EHBOS captures the linear relationship between the two features and detects anomalies as deviations from this dependency. The outliers, which lie far from the expected regression line, are appropriately identified with high anomaly scores by EHBOS. This is evident in the scatter plot of EHBOS scores for this dataset (Figure 4), where the outliers are clearly distinguishable from the normal points.

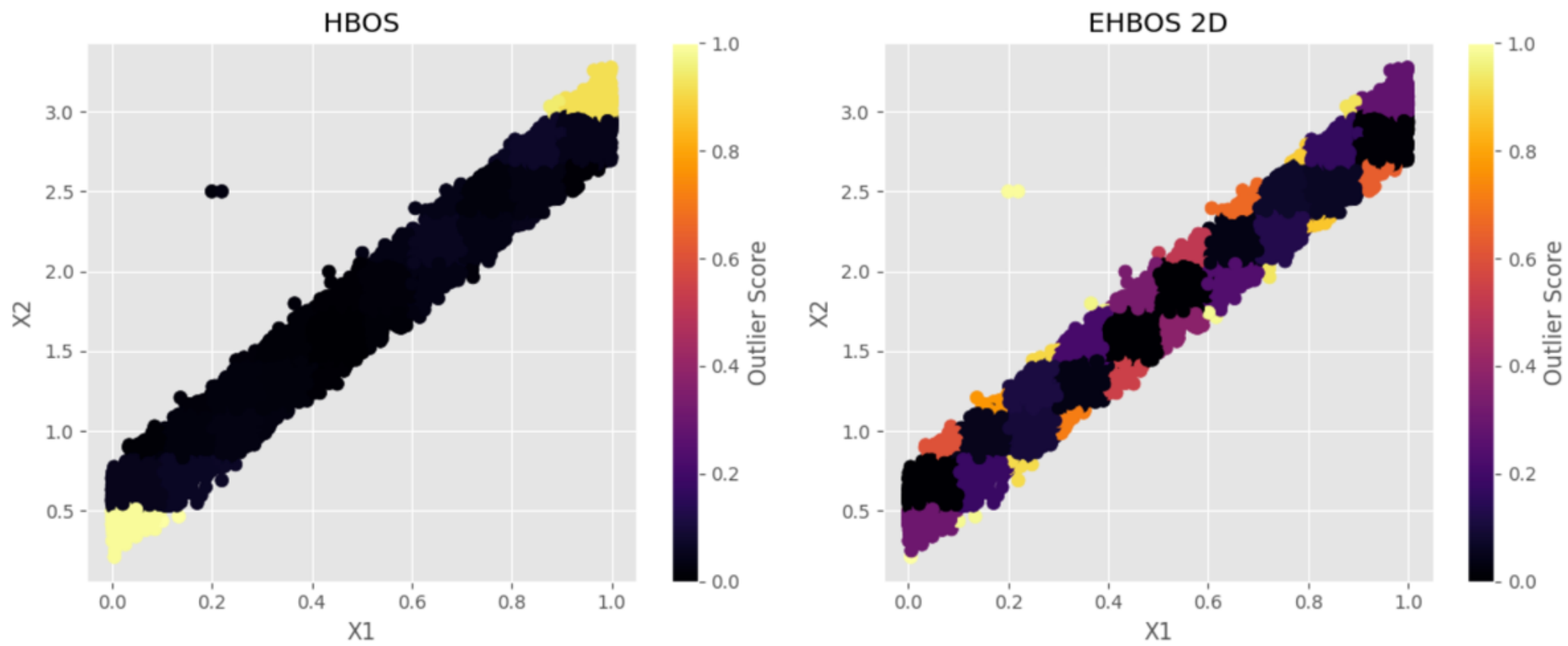

**Figure 4:** Comparison between HBOS and EHBOS (2D) outlier scores for the second example dataset.

These results underscore the importance of modeling feature interactions for effective anomaly detection in datasets where anomalies are defined by deviations from joint feature distributions. By extending HBOS to account for feature dependencies, EHBOS demonstrates superior performance in these scenarios, further validating its design and utility.

### 4.2 Empirical Evaluation

The performance of EHBOS is now evaluated against HBOS on 17 benchmark datasets, encompassing a broad spectrum of anomaly detection challenges. These datasets span various domains, including but not limited to, image, time-series, and tabular data, providing a comprehensive testbed for anomaly detection algorithms. The dataset encompasses different aspects of real-world anomaly detection tasks, such as varying levels of class imbalance, noise, feature correlations, and dimensionality (Table 1).

**Table 1:** 17 benchmark datasets used in the empirical evaluation.

| Data | #Samples | # Dimensions | Outlier Perc |
|---|---|---|---|
| **arrhythmia** | 452 | 274 | 14.6018 |
| **cardio** | 1831 | 21 | 9.6122 |
| **glass** | 214 | 9 | 4.2056 |
| **ionosphere** | 351 | 33 | 35.8974 |
| **letter** | 1600 | 32 | 6.25 |
| **lympho** | 148 | 18 | 4.0541 |
| **mnist** | 7603 | 100 | 9.2069 |
| **musk** | 3062 | 166 | 3.1679 |
| **optdigits** | 5216 | 64 | 2.8758 |
| **pendigits** | 6870 | 16 | 2.2707 |
| **pima** | 768 | 8 | 34.8958 |
| **satellite** | 6435 | 36 | 31.6395 |
| **satimage-2** | 5803 | 36 | 1.2235 |
| **shuttle** | 49097 | 9 | 7.1511 |
| **vertebral** | 240 | 6 | 12.5 |
| **vowels** | 1456 | 12 | 3.4341 |
| **wbc** | 378 | 30 | 5.5556 |

For our empirical evaluations, the number of bins employed in EHBOS's 1D and 2D histogram constructions is adaptively determined for each dataset and feature dimension. This is guided by established statistical principles, known as Birge Rozenholc method (Birgé & Rozenholc, 2006). It ensures an optimal representation of the data's underlying distribution for density estimation. Critically, this same adaptive binning strategy is also applied to HBOS to ensure a fair comparison between the two algorithms, preventing binning differences from skewing the results.

To ensure robustness and fairness in comparison, the evaluation metric used is the Receiver Operating Characteristic (ROC) Area Under the Curve (AUC). This metric is widely used in anomaly detection to measure an algorithm's ability to distinguish between normal and anomalous instances across different decision thresholds. By using the ROC AUC, we account for both the true positive rate (sensitivity) and the false positive rate (1-specificity), providing a comprehensive view of each algorithm's performance across the full range of potential anomaly thresholds.

**Table 2:** Performance comparison of EHBOS and HBOS on 17 benchmark datasets. This table presents the Receiver Operating Characteristic (ROC) Area Under the Curve (AUC) scores for EHBOS and HBOS across 17 diverse benchmark datasets, highlighting each algorithm's effectiveness in distinguishing normal instances from anomalies.

| Data | HBOS ROC AUC | EHBOS ROC AUC |
|---|---|---|
| arrhythmia | 0.8125 | **0.8175** |
| cardio | 0.8511 | **0.9037** |
| glass | 0.7003 | **0.8515** |
| ionosphere | 0.6546 | **0.6666** |
| letter | 0.5903 | **0.6028** |
| lympho | 1.0000 | 0.8556 |
| mnist | 0.6140 | 0.4647 |
| musk | 1.0000 | **1.0000** |
| optdigits | 0.8279 | 0.6267 |
| pendigits | 0.9283 | 0.9238 |
| pima | 0.6956 | 0.6954 |
| satellite | 0.7516 | **0.7886** |
| satimage-2 | 0.9814 | **0.9921** |
| shuttle | 0.9850 | **0.9942** |
| vertebral | 0.3095 | **0.3594** |
| vowels | 0.6807 | **0.7401** |
| wbc | 0.9583 | 0.9533 |

The ROC AUC scores for both HBOS and EHBOS on all 17 benchmark datasets are tabulated in Table 2, providing a clear comparison of the strengths and weaknesses of each algorithm across a diverse set of anomaly detection challenges. The results, summarized in the table, demonstrate that EHBOS outperforms HBOS on several datasets. For example, in datasets like glass (0.87 vs. 0.70), cardio (0.91 vs. 0.85), and satellite (0.79 vs. 0.75), EHBOS achieves a noticeable improvement in ROC AUC. These gains are primarily attributed to EHBOS's ability to capture feature interactions through the inclusion of two-dimensional histograms, a critical enhancement over the univariate feature approach of HBOS. While EHBOS does not universally outperform HBOS, its enhancements are evident in scenarios where feature dependencies significantly influence anomaly detection. For instance, in datasets like glass and cardio, the anomalies are often defined by deviations in joint feature distributions rather than univariate distributions, enabling EHBOS to identify these more effectively. Even in cases where EHBOS underperforms slightly, its scores are competitive with those of HBOS, demonstrating its robustness.

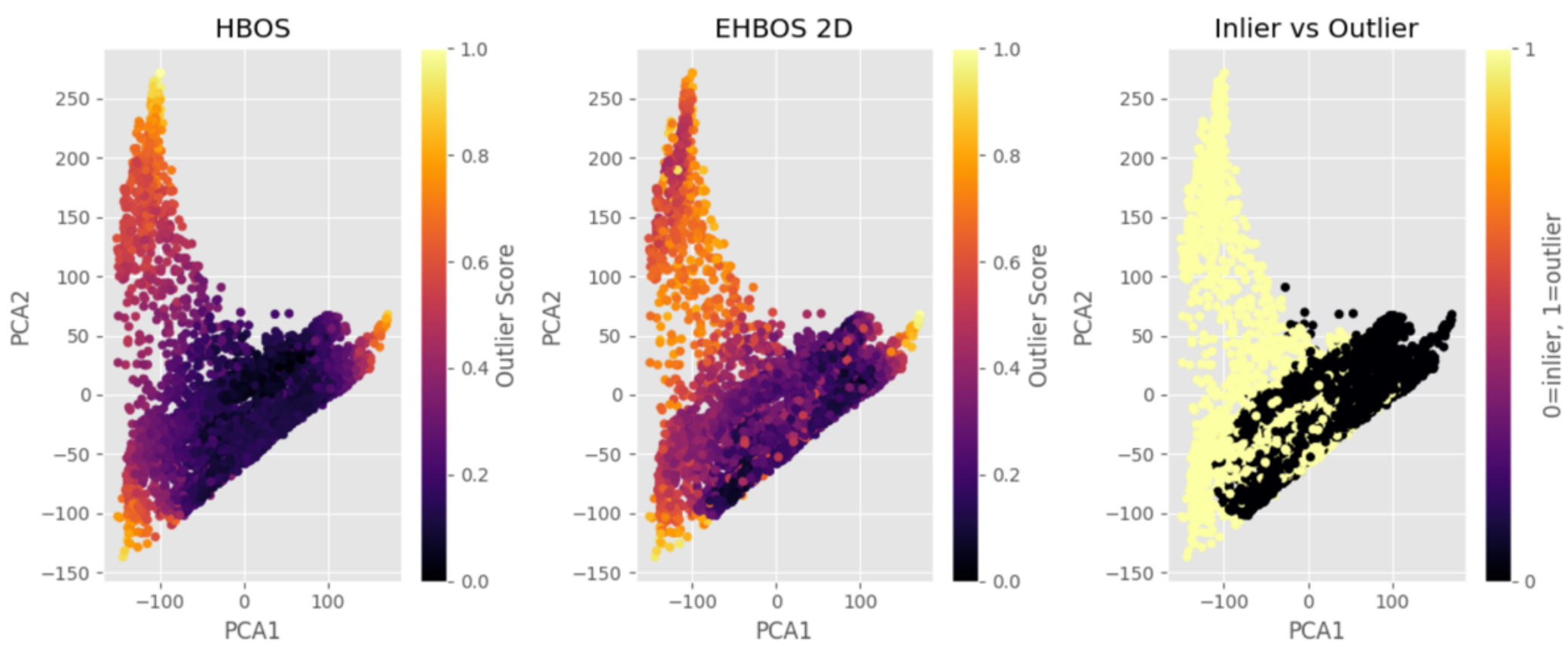

**Figure 5:** Comparison of HBOS and EHBOS outlier scores on the Satellite dataset, reduced to 2D using PCA. The third subplot shows the ground truth, highlighting how EHBOS scores better align with true anomalies.

As an example, we provide a comparative visualization of HBOS and EHBOS (2D) outlier scores in Figure 5 for satellite dataset. Illustrating outlier scores on higher dimensionality dataset is difficult. As such, we apply Principal Component Analysis (PCA) to reduce the satellite dataset to two principal components for visualization. The first subplot displays HBOS scores projected onto the 2D PCA space, while the second shows the EHBOS (2D) scores on the same projection. For reference, the third subplot illustrates the true inliers and outliers based on the dataset's ground truth labels. As evident from the figure, the higher outlier scores assigned by EHBOS align more closely with the true outlier clusters in the satellite dataset compared to the scores generated by HBOS. This visual evidence supports our quantitative findings, highlighting how EHBOS, by incorporating pairwise feature dependencies, can more effectively identify anomalies in real-world scenarios where complex interactions are key indicators of anomalous behavior.

Overall, these results suggest that EHBOS can serve as a valuable tool in scenarios where capturing feature interactions is critical for anomaly detection. By extending the capabilities of HBOS to model dependencies between features, EHBOS offers a flexible approach that can be particularly effective in domains with high-dimensional or complex datasets, such as cybersecurity, fraud detection, and medical diagnostics.

### 4.3 Time Complexity

Finally, we measure the wall-clock running time of both HBOS and EHBOS for the 17 benchmark datasets. We empirically validate our theoretical $O(n \cdot d)$ and $O(n \cdot d^2)$ bounds for HBOS and EHBOS, respectively. This is evaluated under a fixed bin-width setting for both HBOS and EHBOS.

The benchmark dataset has a varied range of sample sizes n and feature dimensions $d$. We plot each set of running time against the theoretical cost metrics $O(n \cdot d)$ (for HBOS) and $O(n \cdot d^2)$ (for EHBOS). As shown in Figure 6, the gray dashed “1:1” fit-lines closely track the observed data points, with HBOS scaling linearly in $O(n \cdot d)$ and EHBOS scaling quadratically in $O(n \cdot d^2)$. These empirical slopes confirm that our implementations achieve the expected time complexities in practice.

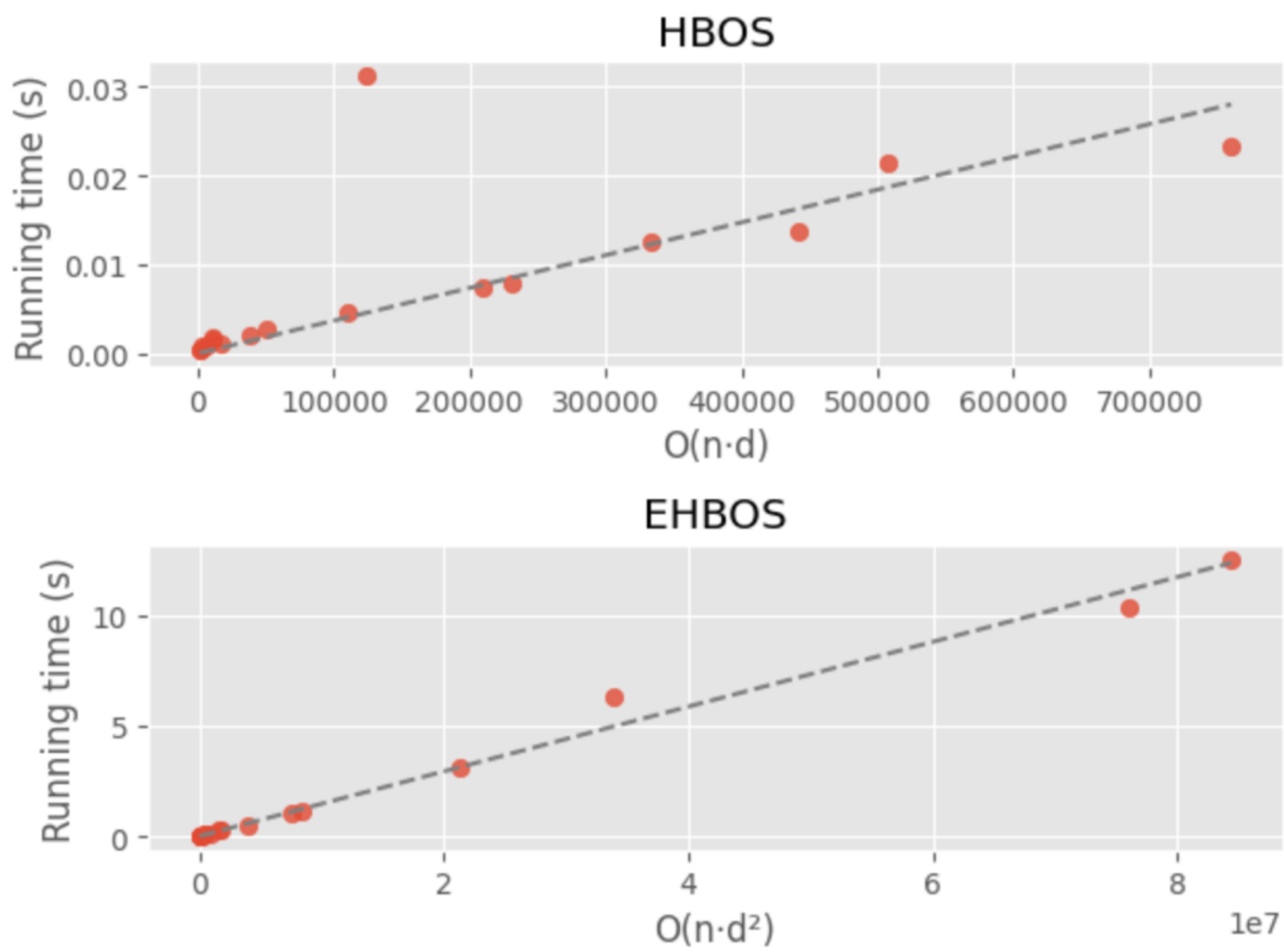

**Figure 6:** Empirical wall-clock runtimes of HBOS (top) and EHBOS (bottom) on all 17 benchmark datasets.

## 5 Conclusion

In this paper, we have introduced the Extended Histogram-Based Outlier Score (EHBOS), a novel extension of the Histogram-Based Outlier Score (HBOS) that incorporates two-dimensional histograms to address the limitations of feature independence inherent in HBOS. By augmenting density estimation with pairwise feature analysis, EHBOS effectively captures feature interactions, enabling improved detection of contextual and dependency-driven anomalies.

Our empirical evaluation across 17 benchmark datasets demonstrates the efficacy of EHBOS, with notable improvements in ROC AUC scores for datasets where feature interactions define the structure of anomalies. Datasets such as glass and cardio highlight the advantages of EHBOS in leveraging joint feature distributions, while competitive performance on datasets dominated by marginal anomalies underscores its robustness. However, the results also reveal that EHBOS is not universally superior, particularly in cases where feature independence aligns with the underlying anomaly patterns. This indicates that the choice between HBOS and EHBOS may depend on the nature of the data and the domain-specific requirements.

EHBOS is generally more appropriate for datasets where anomalies are expected to manifest through complex inter-feature relationships that are not detectable by analyzing features independently. If prior knowledge suggests that feature interactions are crucial for anomaly definition, EHBOS offers a significant advantage. Conversely, for datasets where anomalies are primarily due to extreme values in individual features or where computational efficiency is paramount, HBOS might still be a suitable choice.

EHBOS retains the simplicity of HBOS while extending its applicability to more complex datasets. This makes it a compelling tool for large-scale anomaly detection tasks in domains such as fraud detection, industrial monitoring, and biomedical applications. Future work could focus on optimizing EHBOS for high-dimensional datasets, investigating adaptive methods for feature selection in two-dimensional histograms. The impact of alternative aggregation techniques on EHBOS performance and robustness could also be explored in future work. A full comparative analysis with diverse outlier detection algorithms regarding the efficiency-accuracy trade-off could also be a compelling direction for future work.

By addressing a key limitation of HBOS while maintaining its simplicity and efficiency, EHBOS represents a significant step forward in the development of scalable and interpretable anomaly detection methods. We anticipate that this contribution will inspire further advancements in histogram-based approaches and their applications to diverse real-world problems.

## Conflict of Interest

The author(s) declare no conflict of interest.

## References

Aguilera-Martos, I., García-Barzana, M., García-Gil, D., Carrasco, J., López, D., Luengo, J., & Herrera, F. (2023). Multi-step histogram based outlier scores for unsupervised anomaly detection: ArcelorMittal engineering dataset case of study. *Neurocomputing*, *544*, 126228.

Anbarasi, M., & Dhivya, S. (2017). Fraud detection using outlier predictor in health insurance data. *2017 International Conference on Information Communication and Embedded Systems (ICICES)*, 1–6.

Angiulli, F., & Pizzuti, C. (2002). Fast outlier detection in high dimensional spaces. *European Conference on Principles of Data Mining and Knowledge Discovery*, 15–27.

Aryal, S., Baniya, A. A., & Santosh, K. (2019). Improved histogram-based anomaly detector with the extended principal component features. *arXiv Preprint arXiv:1909.12702*.

Bauder, R. A., & Khoshgoftaar, T. M. (2016). A probabilistic programming approach for outlier detection in healthcare claims. *2016 15th IEEE International Conference on Machine Learning and Applications (ICMLA)*, 347–354.

Benjelloun, F.-Z., Lahcen, A. A., & Belfkih, S. (2019). Outlier detection techniques for big data streams: Focus on cyber security. *International Journal of Internet Technology and Secured Transactions*, *9*(4), 446–474.

Birgé, L., & Rozenholc, Y. (2006). How many bins should be put in a regularhistogram. *ESAIM: Probability and Statistics*, *10*, 24–45.

Breunig, M. M., Kriegel, H.-P., Ng, R. T., & Sander, J. (2000). LOF: identifying density-based local outliers. *Proceedings of the 2000 ACM SIGMOD International Conference on Management of Data*, 93–104.

Chen, Y., Miao, D., & Zhang, H. (2010). Neighborhood outlier detection. *Expert Systems with Applications*, *37*(12), 8745–8749.

Goldstein, M., & Dengel, A. (2012). Histogram-based outlier score (hbos): A fast unsupervised anomaly detection algorithm. *KI-2012: Poster and Demo Track*, *1*, 59–63.

Hilal, W., Gadsden, S. A., & Yawney, J. (2022). Financial fraud: A review of anomaly detection techniques and recent advances. *Expert Systems With Applications*, *193*, 116429.

Hodge, V., & Austin, J. (2004). A survey of outlier detection methodologies. *Artificial Intelligence Review*, *22*, 85–126.

Liu, F. T., Ting, K. M., & Zhou, Z.-H. (2008). Isolation forest. *2008 Eighth Ieee International Conference on Data Mining*, 413–422.

Rousseeuw, P. J., & Hubert, M. (2011). Robust statistics for outlier detection. *Wiley Interdisciplinary Reviews: Data Mining and Knowledge Discovery*, *1*(1), 73–79.

Sakurada, M., & Yairi, T. (2014). Anomaly detection using autoencoders with nonlinear dimensionality reduction. *Proceedings of the MLSDA 2014 2nd Workshop on Machine Learning for Sensory Data Analysis*, 4–11.
Sathe, S., & Aggarwal, C. C. (2018). Subspace histograms for outlier detection in linear time. *Knowledge and Information Systems*, *56*, 691–715.
Yoseph, F., Heikkilä, M., & Howard, D. (2019). Outliers identification model in point-of-sales data using enhanced normal distribution method. *2019 International Conference on Machine Learning and Data Engineering (iCMLDE)*, 72–78.
Zhao, Y., Nasrullah, Z., & Li, Z. (2019). Pyod: A python toolbox for scalable outlier detection. *Journal of Machine Learning Research*, *20*(96), 1–7.